# Combining spatio-temporal appearance descriptors and optical flow for human action recognition in video data

Karla Brkić*, Srđan Rašić*, Axel Pinz‡, Siniša Šegvić* and Zoran Kalafatić*
*University of Zagreb, Faculty of Electrical Engineering and Computing, Unska 3, 10000 Zagreb, Croatia
Email: `karla.brkic@fer.hr`
‡Graz University of Technology, Inffeldgasse 23/II, A-8010 Graz, Austria
Email: `axel.pinz@tugraz.at`

*Abstract*—This paper proposes combining spatio-temporal appearance (STA) descriptors with optical flow for human action recognition. The STA descriptors are local histogram-based descriptors of space-time, suitable for building a partial representation of arbitrary spatio-temporal phenomena. Because of the possibility of iterative refinement, they are interesting in the context of online human action recognition. We investigate the use of dense optical flow as the image function of the STA descriptor for human action recognition, using two different algorithms for computing the flow: the Farnebäck algorithm and the TV-L1 algorithm. We provide a detailed analysis of the influencing optical flow algorithm parameters on the produced optical flow fields. An extensive experimental validation of optical flow-based STA descriptors in human action recognition is performed on the KTH human action dataset. The encouraging experimental results suggest the potential of our approach in online human action recognition.

## I. Introduction

Human action recognition is one of the central topics of interest in computer vision, given its wide applicability in human-computer interfaces (e.g. Kinect [1]), assistive technologies [2] and video surveillance [3]. Although action recognition can be done from static images, the focus of current research is on using video data. Videos offer insight into the dynamics of the observed behavior, but with the price of increased storage and processing requirements. Efficient descriptors that compactly represent the video data of interest are therefore a necessity.

This paper proposes combining the spatio-temporal appearance (STA) descriptors [4] with two variants of dense optical flow for human action recognition. STA descriptors are a family of histogram-based descriptors that efficiently integrate temporal information from frame to frame, producing a fixed-length representation that is refined as new information arrives. As such, they are suitable for building a model of an action online, while the action is still happening, i.e. not all frames of the action are available. The topic of building a refinable action model is under-researched and of great practical importance, especially in video surveillance and assistive technologies, where it is important to raise an alarm as soon as possible if an action is classified as dangerous.

Originally, STA descriptors were built on a per-frame basis, by calculating and histogramming the values of an arbitrary image function (e.g. hue, gradient) over a regular grid within a region of interest in the frame. We propose extending this concept to pairs of frames, so that optical flow computed between pairs of frames is used as an image function whose values are histogrammed within the STA framework. We consider two variants of dense optical flow, the Farnebäck optical flow [5], and the TV-L1 optical flow [6].

## II. Related work

There is a large body of research concerning human action recognition, covered in depth by a considerable number of survey papers (e.g. [7], [8], [9], [10]).

Poppe [8] divides methods for representing human actions into two categories: local and global. In local representation methods, the video containing the action is represented as a collection of mutually independent patches, usually calculated around space-time interest points. In global representation methods, the whole video is used in building the representation. An especially popular category of local representation methods are based on a generalization of the 2D bag-of-visual-words framework [11] to spatio-temporal data. They model local appearance, often using optical flow, and commonly use histogram-based features similar to HOG [12] or SIFT [13]. In this overview we focus on these methods, as they bear the most resemblance to our approach.

The standard processing chain the bag-of-visual-words methods use is summarized in [14]. It includes finding spatio-temporal interest points, extracting local spatio-temporal volumes around these points, representing them as features and using these features in a bag-of-words framework for classification. Interest point detectors and features used are most commonly generalizations of well-known 2D detectors and features. For instance, one of the earliest proposed spatio-temporal interest point detectors, proposed by Laptev and Lindeberg [15], is a spatio-temporal extension of the Harris corner detector [16]. Another example is the Kadir-Brady saliency [17], extended to temporal domain by Oikonomopoulos et al. [18]. Willems et al. [19] introduce a dense scale-invariant spatio-temporal interest point detector, a spatio-temporal counterpart of the Hessian saliency measure. However, there are space-time-specific interest point detectors as well, such as the one proposed by Dollár et al. [20].

Representations of spatio-temporal volumes extracted around interest points are typically histogram-based. For example, Dollár et al. [20] propose three methods of representing volumes as features: (i) by simply flattening the grayscale values in the volume into a vector, (ii) by calculating the histogram of grayscale values in the volume or (iii) by dividing the volume into a number of regions, constructing a local histogram for each region and then concatenating all the histograms. Laptev et al. [21] propose dividing each volume into a regular grid of cuboids using 24 different grid configurations, and representing each cuboid by normalized histograms of oriented gradient and optical flow. Willems et al. [19] build on the work of Bay et al. [22] and represent spatio-temporal volumes using a spatio-temporal generalization of the SURF descriptor. In the generalization, they approximate Gaussian derivatives of the second order with their box filter equivalents in space-time, and use integral video for efficient computation.

Wang et al. [14] present a detailed performance comparison of several human action recognition methods outlined here. The actions are represented using combinations of six different local feature descriptors, and either three different spatio-temporal interest point detectors or dense sampling instead of using interest points. Three datasets are used for the experiments: the KTH action dataset [23], the UCF sports action dataset [24] and the highly complex Hollywood2 action dataset [25]. The best results are 92.1% for the KTH dataset (Harris3D + HOF), 85.6% for the UCF sports dataset (dense sampling + HOG3D), and 47.4% for the Hollywood2 dataset (dense sampling + HOG/HOF). Experiments indicate that regular sampling of space-time outperforms interest point detectors. Also note that histograms of optical flow (HOF) are the best-performing features in 2 of 3 datasets.

The use of STA descriptors [4] can offer a somewhat different perspective on the problem of human action recognition. To build an STA descriptor, one needs a person detector and a tracker, as the STA algorithm assumes that bounding boxes around the object of interest are known. Although this is a shortcoming when compared with the outlined bag-of-visual-words methods that do not need any kind of information about the position of the human, the STA descriptors come with an out-of-the box capability of building descriptions of partial actions, and are therefore worth considering. The bag-of-visual-words methods could also be generalized to support partial actions, but the generalization is not straightforward.

### III. BUILDING STAs WITH OPTICAL FLOW

#### A. Spatio-temporal appearance descriptors

Spatio-temporal appearance (STA) descriptors [4] are fixed-length descriptors that represent a series of temporally related regions of interest at a given point in time. Two variants exist: STA descriptors of the first order (STA1) and STA descriptors of the second order (STA2). Let us assume that we have a series of regions of interest defined as bounding boxes of a human performing an action. The action need not be complete:

assume that the total duration of the action is $T$, and we have seen $t < T$ frames. To build an STA descriptor, one first divides each available region of interest into a regular grid of rectangular patches. The size of the grid, $m \times n$ ($m$ is the number of grid columns and $n$ is the number of grid rows), is a parameter of the algorithm. One then calculates an arbitrary image function (e.g. hue, gradient) for all the pixels of each patch, and represents the distribution of the values of this function over the patch by a $k_1$-bin histogram. Therefore, for each available region of interest, one obtains an $m \times n$ grid of $k_1$-binned histograms, called grid histograms. Let $\mathbf{g}^{(\theta)}$ denote a $m \times n \times k_1$ vector that contains concatenated histogram frequencies of all the $m \times n$ histograms of the grid in time $\theta$, called the grid vector. The STA1 descriptor at time $t$ is obtained by weighted averaging of grid vectors,

$$\mathrm{STA}_1(t) = \sum_{\theta=1}^{t} \alpha_\theta \mathbf{g}^{(\theta)}. \qquad (1)$$

As a weighted average, the STA1 descriptor is an adequate representation of simpler spatio-temporal phenomena, but fails to capture the dynamics of complex behaviors such as human actions. When averaging, the information on the distribution of relative bin frequencies is lost. The STA2 descriptor solves this problem, by explicitly modeling the distribution of each grid histogram bin value over time. Let the vector $\mathbf{c}_i^{(t)}$, called the component vector, be a vector of values of the $i$-th component $\mathbf{g}^{(\theta)}(i)$ of the grid vector $\mathbf{g}^{(\theta)}$ up to and including time $t$, $1 \leq \theta \leq t$:

$$\mathbf{c}_i^{(t)} = \left[\mathbf{g}^{(1)}(i), \mathbf{g}^{(2)}(i), \mathbf{g}^{(3)}(i), \ldots, \mathbf{g}^{(t)}(i)\right]^T. \qquad (2)$$

The STA2 descriptor in time $t$ is obtained by histogramming the $m \times n \times k_1$ component vectors, so that each component vector is represented by a $k_2$-bin histogram, called the STA2 histogram. To obtain the final descriptor, one concatenates the bin frequencies of $m \times n \times k_1$ STA2 histograms into a feature vector,

$$\mathrm{STA}_2(t) = \left[\mathcal{H}_{k_2}(\mathbf{c}_1^{(t)}), \mathcal{H}_{k_2}(\mathbf{c}_2^{(t)}), \ldots, \mathcal{H}_{k_2}(\mathbf{c}_{mnk_1}^{(t)})\right]^T. \qquad (3)$$

Here the notation $\mathcal{H}_{k_2}(\mathbf{c})$ indicates a function that builds a $k_2$-bin histogram of values contained in the vector $\mathbf{c}$ and returns a vector of histogram bin frequencies. Note that the STA1 descriptor has a length of $m \times n \times k_1$ components, while the STA2 descriptor has a length of $m \times n \times k_1 \times k_2$ components.

#### B. Farnebäck and TV-L1 optical flow

We consider using the following two algorithms for estimating dense optical flow: the algorithm of Farnebäck [5] and the TV-L1 algorithm [6].

Farnebäck [5] proposes an algorithm for estimating dense optical flow based on modeling the neighborhoods of each pixel by quadratic polynomials. The idea is to represent the image signal in the neighborhood of each pixel by a 3D surface, and determine optical flow by finding where the surface has moved in the next frame. The optimization is not done on a pixel-level, but rather on a neighborhood-level, so that the optimum displacement is found both for the pixel and its neighbors.

The TV-L1 optical flow of Zach et al [6] is based on a robust formulation of the classical Horn and Schunck approach [26]. It allows for discontinuities in the optical flow field and robustness to image noise. The algorithm efficiently minimizes a functional containing a data term using the L1 norm and a regularization term using the total variation of the flow [27].

*C. Combining STAs and optical flow*

The STA descriptors are built using grids of histograms of values of an arbitrary image function. Optical flow is an image function that considers pairs of images, and assigns a vector to each pixel of the first image. The main issue to be addressed in building spatio-temporal appearance descriptors that use optical flow is how to build histograms of vectors. The issue of building histograms of vectors is well-known, and addressed in e.g. HOG [12] or SIFT [13] descriptors. The idea is to divide the $360°$ interval of possible vector orientations into the desired number of bins, and then count the number of vectors falling into each bin. In HOG and SIFT descriptors, the vote of each vector is additionally weighted by its magnitude, so that vectors with greater magnitudes bear more weight. When building optical flow, it is interesting to consider both the optical flow orientation histograms weighted by magnitude, and the "raw" optical flow orientation histograms, where there is no weighting by magnitude, i.e. all orientation votes are considered equal, as depending on the data the magnitude information can often be noisy.

Note that when combining STA descriptors with optical flow, there is an inevitable lack of flow information for the last frame of the sequence, as there are no subsequent frames. If one wishes to represent the entire sequence, the problem of the lack of flow can be easily solved by building the STA descriptors in the frame before last.

## IV. EXPERIMENTAL EVALUATION

In our experiments, we consider the standard benchmark KTH action dataset and first investigate the effect of optical flow parameters on the produced optical flow and the descriptivity of the derived STA descriptors. Using the findings from these experiments, we then use the best parameters of the flow and optimize the parameters of STA descriptors to arrive at the final performance estimate. Our experiments include only the STA descriptors of the second order (STA2), as STA1 descriptors are of insufficient complexity to capture the dynamics of human actions.

*A. The KTH action dataset*

The KTH human action dataset [23] consists of videos of 25 actors performing six actions in four different scenarios. The actions are walking, jogging, running, boxing, hand waving and hand clapping, and the scenarios are outdoors, outdoors with scale variation, outdoors with different clothes and indoors. The videos are grayscale and of a low resolution ($160 \times 120$ pixels). There is some variation in the viewpoint. The performance of the actions varies among actors, as does the duration. The background is static and homogeneous. A few example frames from the KTH dataset are shown in Fig. 1.

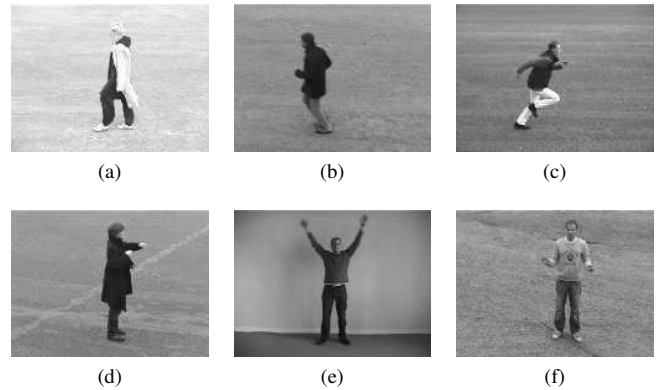

Fig. 1: A few example frames from the KTH dataset: (a) walking, (b) jogging, (c) running, (d) boxing, (e) waving, (f) clapping

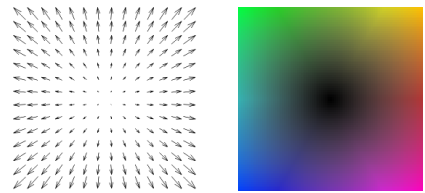

Fig. 2: An illustration of the optical flow coloring scheme that we use. Figure reproduced from [27].

The sequences from the KTH action dataset do not come with a per-frame bounding box annotations. Therefore, in this paper we use the publicly available annotations of Lin et al. [28]. These annotations were obtained automatically, using a HOG detector, and are quite noisy, often only partially enclosing the human.

*B. Optimizing the parameters of optical flow*

In order to compute the optical flow fields necessary for building the STA descriptors, we used the implementations of Farnebäck and TV-L1 optical flow from OpenCV 2.4.5 [29]. Both implementations have a number of parameters that can influence the output optical flow and, in the end, classification performance. To evaluate the influence of individual optical flow parameters on overall descriptivity of the representation, we set up a simple test environment based on a support vector machine (SVM) classifier. We fixed the parameters of the STA descriptor, using a grid of $8 \times 6$ patches, the number of grid histogram bins $k_1 = 8$ and the number of STA2 histogram bins $k_2 = 5$. We then performed a series of experiments where we built a number of STA2 descriptors of the data, varying the values of a single optical flow parameter for each descriptor built. The parameter evaluation procedure can be summarized

as follows:
  (i) For the given parameter set of the optical flow algorithm, calculate the STA2 descriptors over all sequences of the KTH action dataset.
  (ii) Train the SVM classifier using 25-fold cross-validation, so in each iteration the sequences of one person are used for testing, and the sequences of the remaining 24 persons for training.
  (iii) Record cross-validation performance.

We used an OpenCV implementation of a linear SVM classifier, with termination criteria of either $10^5$ iterations or an error tolerance of $10^{-12}$. Other classifier parameters were set to OpenCV defaults, as the goal was not to optimize performance, but to use it as a comparison measure for different optical flow parameters. The optical flow histograms were built both with and without magnitude-based weighting of orientation votes. The obtained results were slightly better when using weighting, so the use of weighting is assumed in all experiments presented here.

*1) Farnebäck optical flow:* For the Farnebäck algorithm, we evaluated the influence of three parameters: the averaging window size $w$, the size of the pixel neighborhood considered when finding polynomial expansion in each pixel $s$, and the standard deviation $\sigma$ of the Gaussian used to smooth derivatives in the polynomial expansion. The remaining parameters were set to their default OpenCV values.

Visualizations of the computed optical flow when the parameters are varied, along with the corresponding obtained SVM recognition rates, are shown in Fig. 3, on an example frame of a running action. In this figure, we adhere to the color scheme for visualization of optical flow proposed by Sanchéz et al. [27], where each amplitude and direction of optical flow is assigned a different color (see Fig. 2). Subfigures 3 (a)-(c) show the influence of the change of parameter $w$, with fixed values of $s$ and $\sigma$, subfigures 3 (d)-(e) show the influence of the change of parameter $s$, with fixed values of $w$ and $\sigma$, and subfigures 3 (g)-(i) show the influence of the change of parameter $\sigma$, with fixed values of $w$ and $s$. It can be seen that the changes along any of the three parameter axes significantly impact the derived optical flow field. Still, the recognition rate in all cases is close to 80%, except when the averaging window size is set to 1, resulting in a recognition rate of 69%. As illustrated in Fig. 3 (a), the optical flow in that case is present only in pixels very close to the human contour, which seems to be insufficient information to build an adequately descriptive representation. Use of a too big averaging window (Fig. 3 (c)) results in noise, and seems to cause a drop in performance. The size of the considered pixel neighborhood does not seem to have a considerable impact on performance, indicating the robustness of the Farnebäck method. Increasing $\sigma$ smooths the derived optical flow, but this blurring lowers performance, so the values of $\sigma$ should be kept low.

*2) TV-L1 optical flow:* For the TV-L1 optical flow, we consider the following parameters: the weight parameter for the data term $\lambda$, the tightness parameter $\theta$, and the time step of the numerical scheme $\tau$.

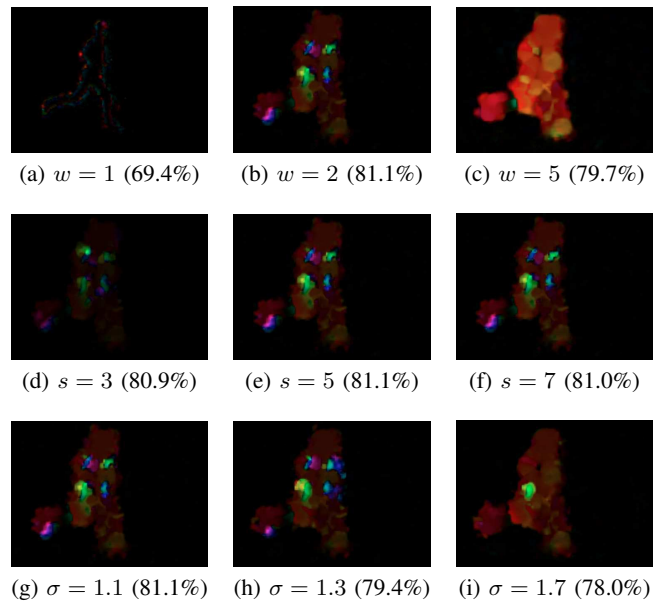

(a) $w = 1$ (69.4%)  (b) $w = 2$ (81.1%)  (c) $w = 5$ (79.7%)

(d) $s = 3$ (80.9%)  (e) $s = 5$ (81.1%)  (f) $s = 7$ (81.0%)

(g) $\sigma = 1.1$ (81.1%)  (h) $\sigma = 1.3$ (79.4%)  (i) $\sigma = 1.7$ (78.0%)

Fig. 3: Obtained Farnebäck optical flow fields and corresponding obtained recognition rates (in parentheses) for varying values of: (a)-(c) parameter $w$ ($s = 5$, $\sigma = 1.1$); (d)-(f) parameter $s$ ($w = 2$, $\sigma = 1.1$); (g)-(i) parameter $\sigma$ ($w = 2$, $s = 5$).

Fig. 4 shows visualizations of the computed optical flow when the parameters are varied, along with the corresponding obtained SVM recognition rates. Subfigures 4 (a)-(c) show the influence of the change of parameter $\lambda$, with fixed values of $\theta$ and $\tau$, subfigures 4 (d)-(e) show the influence of the change of parameter $\theta$, with fixed values of $\lambda$ and $\tau$, and subfigures 4 (g)-(i) show the influence of the change of parameter $\tau$, with fixed values of $\lambda$ and $\theta$. The computed optical flow is noticeably crisper than when using he Farnebäck method, and there is significantly more noise in the background. Setting a mid-range value of the parameter $\lambda$, that influences the smoothness of the derived optical flow, seems to provide the optimum balance between reducing background noise and retaining optical flow information. Parameter $\theta$ should be kept low, as increasing it results in blurring of the derived flow and loss of information on the contour of the human. Parameter $\tau$ does not seem to significantly impact performance, but a mid-range value performs best. Overall, Fig. 4 suggests that STA descriptors benefit from a crisp optical flow around the contour of the human and as low background noise as possible.

### C. Optimizing the parameters of STA descriptors

Based on our analysis of the properties of individual optical flow parameters for the Farnebäck and the TV-L1 algorithm, we were able to select a good parameter combination that should result in solid classification performance. Still, one should also consider optimizing the parameters of the STA descriptors, that were fixed in the previous experiment (grid size of $8 \times 6$, $k_1 = 8$, $k_2 = 5$). Additionally, one should optimize the parameters of the used classifier in order to obtain the optimum performance estimate. We considered simultaneously optimizing STA and classifier parameters with the goal of finding the best cross-validation performance on the KTH action dataset, using STA2 descriptors built using either Farnebäck or TV-L1 optical flow. We fixed the parameters of the Farnebäck and the TV-L1 algorithms to the best-performing ones, as found in the previous section (Farnebäck: $w = 2$, $s = 5$, $\sigma = 1.1$, TV-L1: $\lambda = .05$, $\theta = .1$, $\tau = .15$). A Cartesian product of the following STA parameter values was considered: $m = \{3, 6\}$, $n = \{6, 8\}$, $k_1 = \{4, 5, 8\}$, $k_2 = \{5, 8\}$. For each parameter combination, we performed 25-fold cross-validation multiple times, doing an exhaustive search over classifier parameter space to obtain optimum classifier parameters. Due to heavy computational load involved, we switched from using SVM to using a random forest classifier, because it is faster to train and offers performance that in our experiments turned out to be only slightly reduced when compared to SVM. We optimized the number of trees, the number of features and depth of the random forest classifier. A custom implementation based on the Weka library was used [30]. We repeated the experiments for both Farnebäck and TV-L1-based STA descriptors.

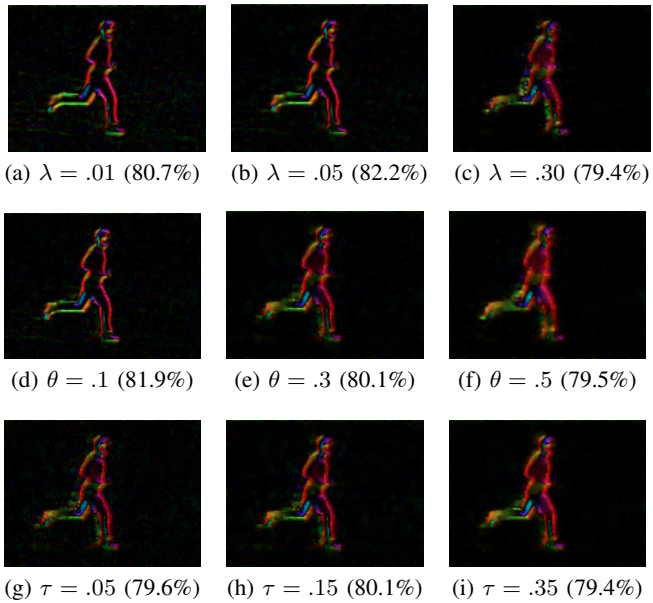

Fig. 4: Obtained TV-L1 optical flow fields and corresponding obtained recognition rates (in parentheses) for varying values of: (a)-(c) $\lambda$ ($\tau = 0.25$, $\theta = 0.3$); (d)-(f) $\theta$ ($\tau = 0.25$, $\lambda = 0.15$); (g)-(i) $\tau$ ($\lambda = 0.15$, $\theta = 0.3$).

|  | Boxing | Clapping | Waving | Jogging | Running | Walking |
|---|---|---|---|---|---|---|
| Boxing | 333 | 27 | 29 | 0 | 7 | 0 |
| Clapping | 34 | 324 | 36 | 0 | 1 | 0 |
| Waving | 28 | 34 | 335 | 0 | 1 | 0 |
| Jogging | 0 | 0 | 0 | 313 | 44 | 43 |
| Running | 2 | 0 | 0 | 84 | 304 | 10 |
| Walking | 1 | 0 | 0 | 22 | 17 | 360 |

TABLE I: The confusion table for the best-performing classifier that uses STA2 feature vectors based on the Farnebäck optical flow to represent KTH action videos. Vertical axis: the correct class label, horizontal axis: distribution over predicted labels.

|  | Boxing | Clapping | Waving | Jogging | Running | Walking |
|---|---|---|---|---|---|---|
| Boxing | 337 | 43 | 9 | 0 | 5 | 2 |
| Clapping | 21 | 318 | 51 | 3 | 3 | 0 |
| Waving | 25 | 51 | 321 | 0 | 1 | 0 |
| Jogging | 7 | 3 | 0 | 287 | 71 | 32 |
| Running | 12 | 4 | 0 | 64 | 318 | 2 |
| Walking | 7 | 0 | 0 | 16 | 6 | 371 |

TABLE II: The confusion table for the best-performing classifier that uses STA2 feature vectors based on the TV-L1 optical flow to represent KTH action videos. Vertical axis: the correct class label, horizontal axis: distribution over predicted labels.

The best recognition rate obtained when using Farnebäck-based descriptors was 82.4%, obtained for an $8 \times 6$ grid, grid histograms of 8 bins and STA2 histograms of 5 bins. The same recognition rate is obtained when using 8 STA2 histogram bins, but we favor shorter representations. The confusion table for the best-performing Farnebäck-based descriptor is shown in Table I. Notice how the confusion centers around two groups: boxing, clapping and waving, and jogging, running and walking. Clapping, waving and boxing are visually similar, as they both include arm movement and static legs. Although in the KTH sequences the boxing action is filmed with the person facing sideways, so the arm movement should occur only on one side of the bounding box, due to noisy annotations it is common that the bounding box of a clapping or a waving action includes only one arm of a person, resulting in a similar motion pattern. The jogging, running, and walking actions are also similar, due to the movement of the legs and the general motion of the body. The greatest confusion is between jogging and running, which is understandable given the variations of performing these actions among actors. Some actors run very similarly to the jogging of others, and vice versa.

For the TV-L1-based STA descriptors, the best obtained recognition rate was 81.6%, obtained for a $3 \times 6$ grid, with 8-bin grid histograms and 5-bin STA2 histograms. As less flow information is generated when using TV-L1, the STA descriptors seem to benefit from using a coarser grid than in the Farnebäck case. The confusion table for the best-performing TV-L1-based descriptor is shown in Table II. Again, the most commonly confused classes can be grouped into two groups: boxing, clapping and waving, and jogging, running and walking. The Farnebäck-based descriptors seem to be better in separating examples among these two groups (e.g. when using Farnebäck-based descriptors no jogging actions got classified as boxing, clapping and waving, while when using TV-L1-based descriptors 10 of them did).

## V. Conclusion

In this paper we proposed an extension of the STA descriptors with optical flow and applied the concept to the problem of human action recognition. A detailed experimental evaluation of two different optical flow algorithms has been provided, with an in-depth study of the properties of individual parameters of each algorithm. We obtained encouraging performance rates, with the descriptors based on the Farnebäck optical flow performing slightly better than the descriptors based on TV-L1. The obtained results suggest that a combination of STA descriptors and optical flow could be used as a feasible representation in a setting that requires partial action models. In [31], similar performance (83.4%) on the KTH action dataset was obtained using gradient-based STA descriptors, and that approach generalized well to a partial action setting. Although better performance rates have been obtained on the KTH dataset [32], our approach is simple, easily extended to other applications, and suitable for building a refinable representation. Therefore, we believe that it merits further investigation.

In future work, we plan to obtain better annotations of humans in video in hopes of improving the overall performance, explore the suitability of optical flow-based STAs to partial action data, and train an SVM classifier with optimized parameters and compare it with the random forest classifier.

## Acknowledgment

This research has been supported by the Research Centre for Advanced Cooperative Systems (EU FP7 #285939).